\documentclass[sigconf]{acmart}
\AtBeginDocument{%
  \providecommand\BibTeX{{%
    \normalfont B\kern-0.5em{\scshape i\kern-0.25em b}\kern-0.8em\TeX}}}

\usepackage{amsmath,bm}

\usepackage{booktabs}
\usepackage{hyperref}
\hypersetup{
    colorlinks=true,
    linkcolor=blue,
    filecolor=magenta,
    urlcolor=cyan,
}
\usepackage{listings}
\usepackage{color}
\usepackage{wasysym}
\usepackage{algorithm}
\usepackage{algorithmic}
\definecolor{dkgreen}{rgb}{0,0.6,0}
\definecolor{gray}{rgb}{0.5,0.5,0.5}
\definecolor{mauve}{rgb}{0.58,0,0.82}
\DeclareMathOperator*{\argmax}{argmax}
\setcopyright{iw3c2w3}
\begin{document}
\setlength{\textfloatsep}{0.2cm}

\title{PyODDS: An End-to-end Outlier Detection System with Automated Machine Learning}

\author{Yuening Li$^{1}$, Daochen Zha$^{1}$, Praveen Kumar Venugopal$^{1}$, Na Zou$^{2}$, Xia Hu$^{1}$}
\email{{liyuening, daochen,zha, razorvine, nzou1, xiahu}@tamu.edu}
\affiliation{%
  \institution{$^{1}$Department of Computer Science and Engineering, Texas A\&M University}
    \institution{$^{2}$Department of Industrial \& Systems Engineering, Texas A\&M University}
}

% \author{Anonymous}
% \affiliation{Affiliation
% }
% \email{anonymous@affiliation.org}

% \affiliation{%

% \affiliation{%

% \affiliation{%

% \affiliation{%





% \vspace{10pt}

\begin{abstract}
Outlier detection is an important task for various data mining applications. Current outlier detection techniques are often manually designed for specific domains, requiring large human efforts of database setup, algorithm selection, and hyper-parameter tuning. To fill this gap, we present \textbf{PyODDS}, an automated end-to-end \textbf{Py}thon system for \textbf{O}utlier \textbf{D}etection with \textbf{D}atabase \textbf{S}upport, which automatically optimizes an outlier detection pipeline for a new data source at hand. Specifically, we define the search space in the outlier detection pipeline, and produce a search strategy within the given search space. PyODDS enables end-to-end executions based on an Apache Spark backend server and a light-weight database. It also provides unified interfaces and visualizations for users with or without data science or machine learning background. In particular, we demonstrate PyODDS on several real-world datasets, with quantification analysis and visualization results.
\end{abstract}

\vspace{-10pt}

\copyrightyear{2020}
\acmYear{2020}
\acmConference[WWW '20 Companion]{Companion Proceedings of the Web Conference 2020}{April 20--24, 2020}{Taipei, Taiwan}
\acmBooktitle{Companion Proceedings of the Web Conference 2020 (WWW '20 Companion), April 20--24, 2020, Taipei, Taiwan}
\acmPrice{}
\acmDOI{10.1145/3366424.3383530}
\acmISBN{978-1-4503-7024-0/20/04}
\maketitle
\settopmatter{printacmref=true}
\section{Introduction}

Outliers refer to the objects with patterns or behaviors that are significantly rare and different from the rest of the majority. Outlier detection plays an important role in various applications, such as fraud detection, cyber security, medical diagnosis, and industrial manufacturer. The research of outlier detection traces far back, and numerous approaches have been proposed to tackle the problem. Representative categories of outlier detection approaches include density-based, distance-based and model-based approaches.

%Overtime, a variety of anomaly detection approaches have been specifically developed for certain application domains.

% Many anomaly detection techniques have been specifically developed
% for certain application domains. Cluster-, distance- and density-based techniques were consequently developed, yet it is hard to find one comprehensive and generic anomaly detection approach to cover all of the cases in the real-world applications.

Despite the exciting results in outlier detection research, it is challenging and expensive to apply outlier detection to tackle real-world problems. First, there is no single outlier detection algorithm outperforms the others on all scenarios, since many outlier detection techniques have been  specifically developed for certain application domains~\cite{li2019specae,li2019deep,huang2019graph}; Second, most outlier detection methods highly depend on their hyper-parameter settings; Third, the contamination ratio of outliers in the given task is usually unknown.

% . given a specific task such as monitoring/alerting systems.

Recently, efforts have been made to integrate various outlier detection algorithms into a single package. Existing approaches~\cite{li2019pyodds,zhao2019pyod} contain different outlier detection methods with various programming languages,  yet they do not tackle with optimal pipeline design as searching and exploration problems, and do not cater specifically to backend-servers for large-scale applications.

In the meanwhile, a large focus of the machine learning community has been to find better hyper-parameter settings, which has been successfully tackled using Bayesian optimization, reinforcement learning, etc., and forms a core component of AutoML systems. However, less attention has been paid to finding a good solution for an end-to-end, joint optimization problem including multiple components, especially in real-world data mining tasks.

% Thus, these packages still require large efforts of database setup, algorithm selection, and hyperparmeter tuning.
% These toolkits greatly facilitate the applications of outlier detection algorithms, yet they are mainly focus on static data.

%It would be necessary to design a systematic strategy to autonomously figure out the most appropriate approach and model architecture to achieve optimal solution for the outlier detection task. In this case, we may need to explore good model architecture to extract informative features, as well as to figure out the most effective detection algorithm.

% It is a challenging task to develop the above unified outlier detection system. First, the burden of throughput is heavier than usual since time window based approaches require to access and query the data in a more frequent way. Second, loading data from remote servers causes a large cost of moving data outside database server or over the network for analysis. Third, the system should support static and time series data with unified API.

To bridge the gap, we present \textbf{PyODDS}, a full-stack, end-to-end system for outlier detection. $\textbf{PyODDS}$ has desirable features from the following perspectives. First, to our best knowledge, PyODDS describes the first attempt to incorporate automated machine learning with outlier detection, and belongs to one of the first attempts to extend automated machine learning concepts into real-world data mining tasks. Second, we carefully design an end-to-end framework for outlier detection, including database operations and maintenance, the search process of automated outlier detection~(including the search space and the search strategy design). Finally, we present a visual analytic system based on our proposed framework for demonstration.

% PyODDS designs a unified API with detailed documentation, such as outlier detection approaches, database operations, and visualization functions for demonstrations.

% First, it contains 13 algorithms, including statistical approaches, and recent neural network frameworks. Second, PyODDS supports both static and time series data analysis, with flexible time-slices segmentation. Third,  PyODDS supports operation and maintenance from a light-weight SQL based database, which reduces the cost of queries and loading data from different remote servers. Fourth, PyODDS provides visualization tools for the original distribution of raw data, and predicted results, which offers users a direct and vivid perception. Last, PyODDS includes a unified API with detailed documentation, such as outlier detection approaches, database operations, and visualization functions.

\begin{figure*}[t]
  \includegraphics[width=0.95\linewidth]{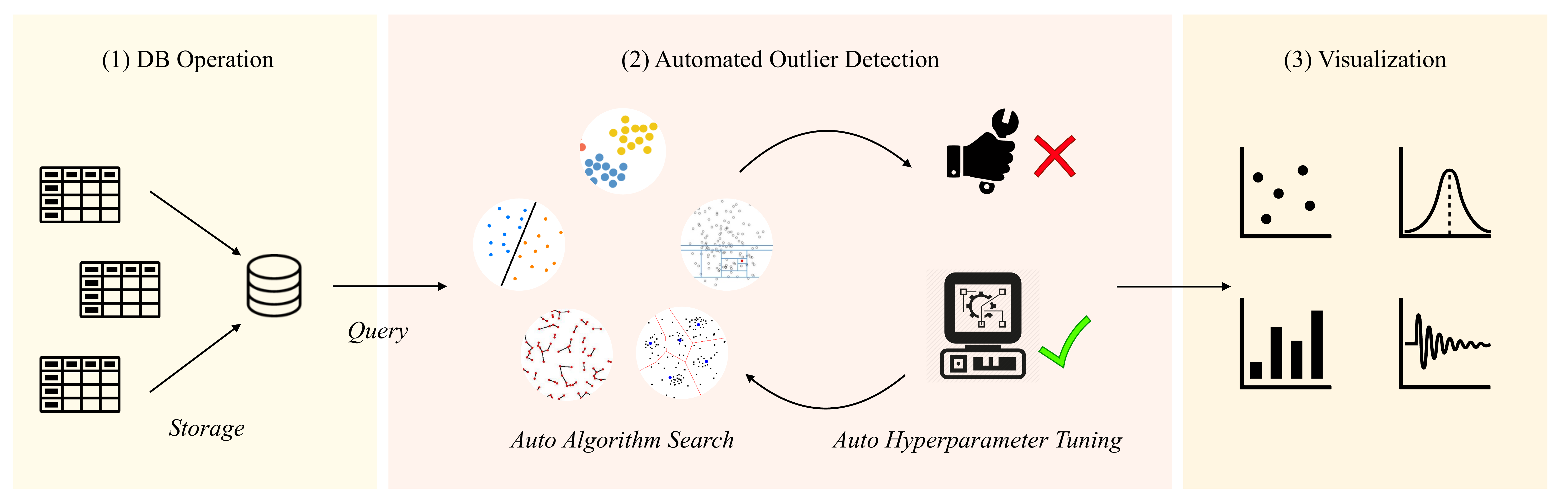}
  \vspace{-8pt}
  \caption{Overview of PyODDS}
  \label{pipeline}
\end{figure*}
  \vspace{-8pt}

\section{PyODDS System Architecture}

The pipeline from query data to evaluation and visualization is outlined in Figure ~\ref{pipeline}, which consists of 3 components. The first component is the information extraction, which collects the source data via {\verb|query|} functions with flexible time-slices segmentation, including user-info confirmation, database operation, and maintenance. The second component is the suspicious outlier detection. It detects suspicious instances with traditional outlier detection approaches as an automated machine learning problem, including the search space design and  the search strategy development. The last component is the visualization part, which designs for users to understand the detection results better. In the following subsections, we focus on the first and second components. We will discuss the visualization in Section 4.

\subsection{Information Extraction}

In this component, we extract the information from a specific time range through database operations. PyODDS includes database operation functions for client users: (1) $connect\_server$ function allows the client  to connect the server with host address and user information for safety verification; (2) $query\_data$ function designs for  flexible time-slices segmentation.

\subsection{Automated Outlier Detection}

To detect suspicious outliers, we need to find the best pipeline configuration. We formulate the problem of finding the best policy as a conjunctive  search problem. In this component, our method consists of two subsections: a search space and a search strategy.

\subsubsection{Search Space}

In our search space, a policy consists of sub-policies as a batch of outlier detection algorithms. Additionally, the policy also contains the hyper-parameters as another conditional sub-policy: 1) specific hyper-parameter settings corresponding to each algorithm sub-policy  which controls the learning process; 2) the contamination ratio which determines the portion of outliers  corresponding to the given data source.

Each algorithm we included also comes with a default range of hyper-parameter settings. Within each algorithm sub-policy, hyper-parameters which might be discrete, ordinal, or continuous, need to be optimized in the meanwhile.

% The searchwhich will be described in more detail in the following section.

\subsubsection{Search Strategy}

Following the search space setting we proposed above, we define the problem of automated outlier detection with algorithm selection and hyper-parameter tuning as follows. Let $\mathcal{A}=\{A_1, A_2,..., A_N\}$ be a set of outlier detection algorithms, and $\bm{\lambda}=\{\lambda_1, \lambda_2,..., \lambda_N\}$ be the set of corresponding hyper-parameters. We assume $\bm{\lambda}$ is given. Let $\mathcal{D}^{train}$ and $\mathcal{D}^{val}$ be the training set and validation set, respectively. Denote $\mathcal{M}({A}^s, \bm{\lambda}^s, \mathcal{D}^{train}, \mathcal{D}^{val})$ as the performance on $\mathcal{D}^{val}$ in terms of metric $M$ when trained on $D^{train}$ with algorithm  ${A}^s \subseteq \mathcal{A}$ and corresponding hyper-parameters $\bm{\lambda}^s \subseteq \bm{\lambda}$. The algorithm is to find optimal solution ${A}^*, \bm{\lambda}^*$ via observation history $\mathcal{H}$. We define the objective as
\begin{equation}
    {A}^*, \bm{\lambda}^* \in \argmax_{{A}^s \subseteq \mathcal{A}, \bm{\lambda}^s \subseteq \bm{\lambda}^*} M({A}^s, \bm{\lambda}^s, \mathcal{D}^{train}, \mathcal{D}^{val}).
    \label{eqn:autodef}
\end{equation}

\begin{algorithm}[thb]
\caption{Optimization Process}\label{algo:optimization}
\setlength{\belowcaptionskip}{-5cm}
\begin{algorithmic}[1]
   \STATE {\bfseries Input:} $\mathcal{H}, \mathcal{A}, \bm{\lambda}, \mathcal{D}^{train}, \mathcal{D}^{val}$, $T_{max}$
   % \STATE {\bfseries Output:} $f$, $O$
\STATE $T\leftarrow 1$, \
\WHILE { $T < T_{max}$ }
  \STATE $T\leftarrow T + 1$
  \STATE ${A}^{t}$ $\leftarrow$ $M(\mathcal{A}^{t-1}, \bm{\lambda}^{t-1}, \mathcal{D}^{train}, \mathcal{D}^{val})$\
  \STATE $\lambda^{t}_{\mathrm{discrete}}$ $\leftarrow$ $M(\mathcal{A}^{t}, \bm{\lambda}^{t-1}, \mathcal{D}^{train}, \mathcal{D}^{val})$\
  \STATE $\lambda^{t}_{\mathrm{continuous}}$ $\leftarrow$ $M(\mathcal{A}^{t}, \bm{\lambda}^{t-1}, \mathcal{D}^{train}, \mathcal{D}^{val})$\
  \STATE $\mathcal{H}$ $\leftarrow$ $\mathcal{H}$ $\cup$ \{$\mathcal{A}^{t}$, $\lambda^{t}$\}
%   \FOR{$o \in \Omega(f)$}
%     \STATE $f'\leftarrow\mathcal{M}(f, \{o\})$
%     \IF {$e^{\frac{c_{min} - \alpha(f')}{T}} > Rand()$}
%       \STATE $Q$.Push($f'$)
%     \ENDIF
%     \IF {$ c_{min} >\alpha(f')$}
%       \STATE $c_{min} \leftarrow \alpha(f')$, $f_{min} \leftarrow f'$
%     \ENDIF
%   \ENDFOR
\ENDWHILE

\STATE {\bfseries Return}  ${A}^*, \lambda^*$  with the best performance in $\mathcal{H}$
\end{algorithmic}
\end{algorithm}
% \vspace{-2pt}

To get a step further, Sequential Model-Based Global Optimization (SMBO) algorithms have been used in many applications where evaluation of the fitness function is expensive, i.e., automated machine learning tasks~\cite{bergstra2011algorithms}. To optimize the evaluation function $\mathcal{M}(\mathcal{A}^s, \bm{\lambda}^s, \mathcal{D}^{train}, \mathcal{D}^{val})$, we optimize the criterion of Expected Improvement, the expectation under $\mathcal{A}^s, \bm{\lambda}^s$ when $y=\mathcal{M}(\mathcal{A}^s, \bm{\lambda}^s,\\ \mathcal{D}^{train}, \mathcal{D}^{val})$ negatively exceed the threshold $y^{*}$:
\begin{equation}
    \mathrm{EI}_{y^{*}}(x)=\int_{-\infty}^{y^{*}}\left(y^{*}-y\right) p(y | x) d y=\int_{-\infty}^{y^{*}}\left(y^{*}-y\right) \frac{p(x | y) p(y)}{p(x)} d y,
\end{equation}
where the  point $x^{*}$ that maximizes the surrogate (or its transformation) becomes the proposal for where the  function should be evaluated.

The tree-structured Parzen estimator (TPE) models $p(x|y)$ by transforming to a generative process, which replaces the distributions of the configuration prior to non-parametric densities. We borrow the strategy in ~\cite{bergstra2011algorithms} here to minimize the EI. We keep the Estimation of Distribution (EDA, ~\cite{larranaga2001estimation}) approach on the discrete part of our search space (algorithm selection and discrete hyper-parameters), where we sample candidate points according to binomial distributions, while we use the Covariance Matrix Adaptation - Evolution Strategy, a gradient-free evolutionary algorithm (CMA-ES, ~\cite{hansen2006cma}) for the remaining part of our search space (continuous hyper-parameters). The whole optimization process can be summarized in Algorithm~\ref{algo:optimization}.

\begin{figure*}[t]
  \includegraphics[width=\linewidth]{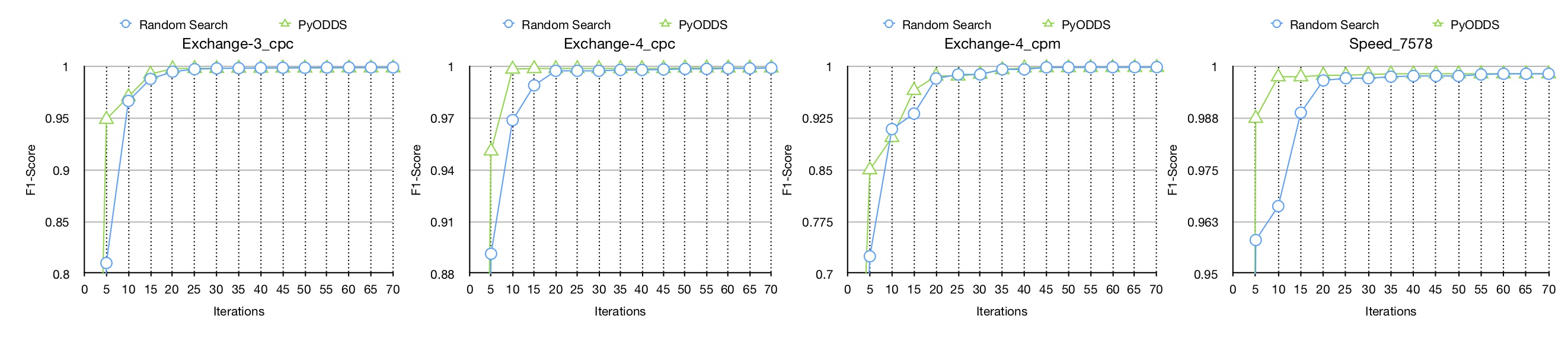}
      \vspace{-20pt}
  \caption{Progression of top-5 averaged performance of different search methods, i.e., Random Search and PyODDS.}
  \label{fig:topk}
    \vspace{-10pt}
\end{figure*}

\section{Experimental Evaluation}

\subsection{Data Source}
The time-series data, which is used to train and evaluate PyODDS, comes from a benchmark dataset, NAB corpus~\cite{ahmad2017unsupervised}. NAB corpus contains 58 different individual tasks with ground-truth. The reasons why we employ this data
source are in three folds. First, NAB corpus provides fine-grained labels, where the core
principles in independence, transparency and fairness guarantee. Second, the data in NAB corpus are ordered, timestamped, which cover a varies of real-world application scenarios, including server monitor logs from AmazonCloudwatch service, online advertisement clicking-rates, real-time traffic transportation, and collection from Twitters with trading related contents. Third, each raw data file is a dictionary of key-value pair, which is naturally to be represented as tabular data that meets the requirements of the backend database service in the PyODDS.

\subsection{Algorithm Space Configurations}
We implemented 13 state-of-the-arts outlier detection algorithms as the search space, including statistical approaches, and recent neural network frameworks. In the meanwhile, in order to support both static and time series data analysis, the search space covers algorithms with different settings.

\subsection{Detection Results Evaluation}

In this section, we empirically investigate the performance of PyODDS to answer the following questions: first, how does the algorithm with hyper-parameters discovered by PyODDS compare with state-of-the-art handcrafted algorithms? Second, how does the search process affect performance?

In Table~\ref{tab:freq}, we show the performance on the NAB corpus. We follow the default setting in NAB as the scoring algorithm, which uses a scaled sigmoidal scoring function to quantify the detection performance. The smooth score function  ensures that small labeling errors will not cause large changes in reported scores. The evaluation matrix includes the standard profile, reward low FPs, and reward low FNs. The standard profile
assigns TPs, FPs, and FNs with relative weights, and the latter two profiles accredit greater penalties for FPs and FNs,
respectively. For more detailed definitions, please refer to the default setting~\cite{lavin2015evaluating}.

To answer the first question, we use PyODDS to find the best policies on the NAB corpus. As can be seen from Table~\ref{tab:freq}, the outlier detection solution discovered by PyODDS architecture achieves competitive performance with current
state-of-the-art models: the handcrafted algorithms, and random searched results. It shows that PyODDS could find optimal solutions within a large range of configurations for different detection tasks.

\begin{figure*}[t]
   \includegraphics[width=0.78\linewidth]{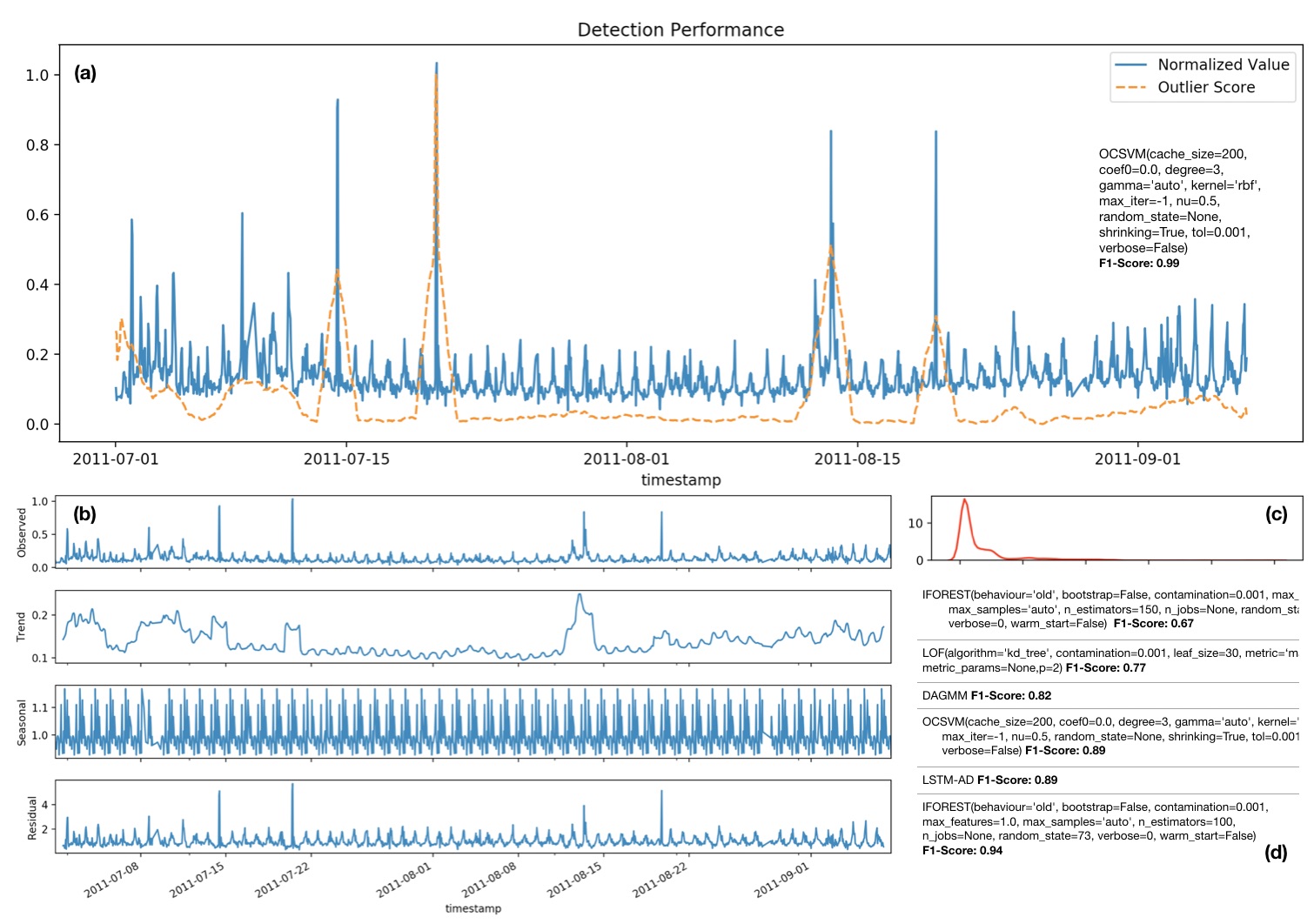}
      \vspace{-10pt}
  \caption{Demonstration of using PyODDS in visualizing prediction result}
  \label{fig:visualization}
    \vspace{-10pt}
\end{figure*}

For the second question, we conduct the search process in the same search space with different search strategies. As can be seen from Figure~\ref{fig:topk}, PyODDS is more efficient in finding
the well-performed architectures during the search progress. Comparing with the random search, the top-5 architectures discovered by PyODDS have better performance (F1-score) and could convergence faster on different datasets. It shows the effectiveness of the search strategy PyODDS implemented could enhance the performance and accelerate the search efficiency.

\vspace{-10pt}

\begin{table}
\small
\resizebox{0.48\textwidth}{!}{
  \begin{tabular}{c|ccc}
    \toprule
    Model & Standard Profile&Reward Low FP&Reward Low FN\\
    \midrule
    Perfect & 100 & 100 & 100\\
      \midrule
     CBLOF~\cite{he2003discovering}  & 94.56 & 93.29 & 96.68\\
     HBOS~\cite{aggarwal2015outlier}  & 91.86 & 95.74 & 93.47\\
     IFOREST~\cite{liu2008isolation}  & 92.44 & 92.10 & 94.38\\
     KNN~\cite{ramaswamy2000efficient}  & 90.76 & 96.12 & 93.42\\
     LOF~\cite{breunig2000lof}  & 92.61 & 88.78 & 89.86\\
     OCSVM~\cite{scholkopf2001estimating}  & 88.63 & 94.60 & 91.31\\
     PCA~\cite{shyu2003novel}   & 93.15 & 94.50 & 96.28\\
     RobustCovariance  & 96.68 & 95.27 & 94.76\\
     SOD~\cite{kriegel2009outlier}  & 78.46 & 78.46 & 82.93\\
     AUTOENCODER~\cite{hawkins2002outlier}  & 94.74 & 96.41 & 93.64\\
     DAGMM~\cite{zong2018deep}  & 85.27 & 83.35 & 90.21\\
     LSTMAD~\cite{malhotra2015long}  & 93.19 & 95.18 & 92.43\\
     LSTMENCDOC~\cite{malhotra2016lstm}  & 94.31 & 89.23 & 89.23\\
     %LUMINOL  & 53.3 & 53.1 & 68.3\\
     \midrule
    RANDOM   & 87.38 & 90.79 & 86.90\\
    PyODDS & 96.68 & 95.27 & 94.76\\

    % PyODDS  & 52.11 & 51.09 &  67.50\\

  \bottomrule
\end{tabular}}
  \caption{Test performance comparison for outlier detection algorithms: the state-of-the-art algorithms, the optimal solution found by random search baseline, the optimal solution found by PyODDS.}
  \vspace{-18pt}
  \label{tab:freq}
\end{table}

\section{Demonstration}

PyODDS is composed of a frontend and a server backend. Our system is written in Python and uses Apache Spark as the server backend and TDengine as the database support service. We demonstrate our system based on the real-world datasets from the Numenta Anomaly Benchmark (NAB) corpus~\cite{ahmad2017unsupervised}.

First, after selecting the data source and time range, our system will automatically find an algorithm with default hyper-parameter settings from the search space, and show the detection results. Illustrated by Figure~\ref{fig:visualization}(a), we provide the normalized value from the original time series as the blue line, and outlier score as orange line, to help users understand the data distribution in the original data source, as well as the detection results. Lower outlier score indicates that the data point is considered ``normal''. Higher values indicate the presence of an outlier in the data.

% Historical records could help user to better understand the search process.

In addition, PyODDS provides time series analysis tools for users to better understand the data source. Illustrated in Figure~\ref{fig:visualization}(b), PyODDS decomposes the original time series as a combination of level, trend, seasonality, and residual components. The residual values could also act as denoters for outlier detection in time-series. In the meanwhile, in Figure~\ref{fig:visualization}(c), PyODDS estimates the probability density function of the values in each timestamp, which provides a comprehensive scope of the data distribution in the original data source.  According to the search strategy and search space we proposed in the previous sections, we also provide trace logs to illustrate the search process for records. After several iterations of the search process, the selected algorithms with specific configurations are listed in Figure~\ref{fig:visualization}(d)).  As shown in the user case, extreme values and spikes without seasonal patterns (i.e., in the time stamp 2011-07-15, etc) have larger outlier score than the rest majority as normal cases (shown in (a) and (c)), as well as larger residual value after time series decomposition (shown in (b)). Current best solution is the sub-policy OCSVM with specific hyperparameter settings.

\vspace{-8pt}
\section{Conclusion}

In this demo, we propose an end-to-end approach to detect outliers, and demonstrate the prediction results for users to better understand the data source.  PyODDS automatically search an optimal outlier detection pipeline for a new dataset at hand out of a defined proposed search space via the proposed search strategy.

% $\mathcal{H}=\left(x_{i}, f\left(x_{i}\right)\right)_{i=1}^{n}$

% \begin{equation}
%     \mathrm{EI}_{y^{*}}(x)=\int_{-\infty}^{y^{*}}\left(y^{*}-y\right) p(y | x) d y=\int_{-\infty}^{y^{*}}\left(y^{*}-y\right) \frac{p(x | y) p(y)}{p(x)} d y
% \end{equation}

% \begin{equation}
%     \gamma=p\left(y<y^{*}\right) \text { and } p(x)=\int_{\mathbb{R}} p(x | y) p(y) d y=\gamma \ell(x)+(1-\gamma) g(x)
% \end{equation}

% \begin{equation}
% \begin{split}
%     & \int_{-\infty}^{y^{*}}\left(y^{*}-y\right) p(x | y) p(y) d y \\
%      = & \ell(x) \int_{-\infty}^{y^{*}}\left(y^{*}-y\right) p(y) d y \\
%      = & \gamma y^{*} \ell(x)-\ell(x) \int_{-\infty}^{y^{*}} p(y) d y
% \end{split}
% \end{equation}

% \begin{equation}
%     \begin{split}
%         E I_{y^{*}}(x)=\frac{\gamma y^{*} \ell(x)-\ell(x) \int_{-\infty}^{y^{*}} p(y) d y}{\gamma \ell(x)+(1-\gamma) g(x)} \propto\left(\gamma+\frac{g(x)}{\ell(x)}(1-\gamma)\right)^{-1}
%     \end{split}
% \end{equation}

% \begin{equation}
%     p(x | y)=\left\{\begin{array}{ll}{\ell(x)} & {\text { if } y<y^{*}} \\ {g(x)} & {\text { if } y \geq y^{*}}\end{array}\right.
% \end{equation}

% \begin{equation}
%     \mathrm{EI}_{y^{*}}(x):=\int_{-\infty}^{\infty} \max \left(y^{*}-y, 0\right) p_{M}(y | x) d y
% \end{equation}

\bibliographystyle{ACM-Reference-Format}
\bibliography{main}


\end{document}